%% file: main.tex
\documentclass[runningheads]{llncs}
\usepackage[T1]{fontenc}
\usepackage{graphicx}
\usepackage{booktabs}
\usepackage[misc]{ifsym}

\usepackage{hyperref}       
\usepackage{url}            
\usepackage{amsfonts}       
\usepackage{nicefrac}       
\usepackage{microtype}      
\usepackage{xcolor}         
\usepackage{algorithm}
\usepackage{algorithmic}
\usepackage{mathtools}
\usepackage{dsfont} 
\usepackage{cleveref}

\usepackage{snaptodo}
\snaptodoset{block rise=2em}
\snaptodoset{margin block/.style={font=\tiny}}

\usepackage[normalem]{ulem}

\usepackage{wrapfig} 
\newcommand{\as}[1]{}
\newcommand{\es}[1]{}
\newcommand{\emmm}[1]{}

\usepackage{subcaption}
\usepackage{multirow}

\begin{document}
\title{Harnessing Mixed Features for Imbalance Data Oversampling: Application to Bank Customers Scoring} 

\titlerunning{Harnessing Mixed Features for Imbalance Data Oversampling}

\author{Abdoulaye SAKHO\inst{1,2} \and Emmanuel MALHERBE\inst{1} \and Carl-Erik GAUTHIER\inst{3} \and Erwan SCORNET\inst{2}}

\authorrunning{A. SAKHO \& E. MALHERBE \& C-E. GAUTHIER \& E. SCORNET}
\institute{
Artefact Research Center, Paris, France \email{\{abdoulaye.sakho,emmanuel.malherbe\}@artefact.com} 
\and 
Laboratoire de Probabilités, Statistique et Modélisation Sorbonne Université and Université Paris Cité, CNRS, F-75005, Paris \email{\{erwan.scornet\}@polytechnique.edu}
\and
Société Générale, Paris, France \email{\{carl-erik.gauthier\}@socgen.com}
}

\maketitle              


\begin{abstract}
This study investigates rare event detection on tabular data within binary classification. Standard techniques to handle class imbalance include SMOTE, which generates synthetic samples from the minority class. However, SMOTE is intrinsically designed for continuous input variables. 
In fact, despite SMOTE-NC---its default extension to handle mixed features (continuous and categorical variables)---very few works propose procedures to synthesize mixed features. 
On the other hand, many real-world classification tasks, such as in banking sector, deal with mixed features, which have a significant impact on predictive performances.
To this purpose, we introduce MGS-GRF, an oversampling strategy designed for mixed features. This method uses a kernel density estimator with locally estimated full-rank covariances to generate  continuous features, while categorical ones are drawn from the original samples through a generalized random forest. 
Empirically, contrary to SMOTE-NC, we show that MGS-GRF exhibits two important properties: $(i)$ the coherence i.e. the ability to only generate combinations of categorical features that are already present in the original dataset and $(ii)$ association, i.e. the ability to preserve  the dependence between continuous and categorical features. 
We also evaluate the predictive performances of LightGBM classifiers trained on data sets, augmented with synthetic samples from various strategies.
Our comparison is performed on simulated and public real-world data sets, as well as on a private data set from a leading financial institution.
We observe that synthetic procedures that have the properties of coherence and association display better predictive performances in terms of various predictive metrics (PR and ROC AUC...), with MGS-GRF being the best one.  
Furthermore, our method exhibits promising results for the private banking application, with development pipeline being compliant with regulatory constraints. 
\end{abstract}

\input{introduction}

\input{related_work}
\input{algorithm}

\input{illustrations}

\input{experiments}

\input{conclusion}

\begin{credits}
\subsubsection{\ackname} We would like to express our gratitude to the following individuals for their valuable help and feedbacks : Vincent AURIAU, Mohamed CHTIBA, Jean-Baptiste JANVIER and Martin KLIEBER.

\end{credits}
%
%
\bibliographystyle{splncs04}
\bibliography{refs}
%

\input{appendix}

\end{document}

%% file: introduction.tex
\section{Introduction}
Addressing class imbalance in binary classification presents a significant challenge across various machine learning applications \cite{sun2009classification,spelmen2018review}, such as medical diagnosis, customer churn prediction or anomaly detection \cite{RF-medical-example,nguyen2021comparison,sun2009classification}.
In particular, detecting fraud is a prime issue in banking \cite{ex-fraud,li2024sefraud}: the vast majority of customers make legitimate transactions, while the fraudulent ones represent only a minority but have a significant operational, regulatory, and reputational impact.

Several seminal works have introduced rebalancing strategies in order to improve the predictive performances of generic classifiers \cite{chawla2002smote,mukherjee2021smote}. These strategies can be divided into two categories \cite{sakhowe}: model-level strategies, that aim at adapting an existing algorithm, for example by weighting classes or minimizing a specific loss function \cite{cao2019learning,lin2017focal}; and data-level strategies, that act on the original data set by oversampling or undersampling the observations, and are thus model agnostic. 
A subclass of data-level strategies, named synthetic procedures, generates new samples in the minority class, with many variants introduced in the literature \cite{chawla2002smote,he2008adasyn,han2005borderline}. One key characteristic is that most of them are primarily designed to handle numerical features and thus do not handle categorical features \cite{fernandez2018learning,spelmen2018review}.
%
%
In practice, categorical features are very common in tabular data (e.g. job category, country, gender), and can represent a relevant signal for improving the predictive performances of the learning tasks, such as those described in \cite{grinsztajn2022tree,garchery2018influence}. This highlights the importance of handling mixed features when generating samples for imbalance data. Besides, combination of several categorical variables need to be generated in a coherent way with each other and with respect to the continuous variables.
In this paper, we focus on synthetic rebalancing strategies for tabular data, with the main objective to handle mixed features.
We emphasize that when generating categorical features, a major aspect is to ensure their intrinsic coherence and their association with continuous features. 
The notion of coherence aims at expressing that a combination of categorical variables can be judged as plausible by some business owner, such as a bank analyst. Indeed, generating samples that do not seem credible may lead to a reluctance to apply subsequent machine learning analyses, without mentioning the potential negative impact on the predictive performances. 
We define formally coherent combinations as the ones existing in the original samples.
On the other hand, the association level measures the dependence between continuous and categorical features, via the accuracy of a given model trained to predict the categorical variables based on continuous variables. Thus, preserving the level of association between original and synthetic data ensures that the distribution of categorical variables conditional on continuous variables remains similar. An augmented data set that is coherent and preserves the association level compared to the original data has a distribution close to the original data set. 
Our main contributions are: 
\begin{itemize} 
    \item We introduce MGS-GRF, a strategy for mixed features, with a kernel density estimation for continuous ones and a generalized forest for categorical ones.
    \item On simulated data, we prove that SMOTE-NC, probably the most widely used synthetic rebalancing strategy, is not coherent and does not preserve association, thus creating unplausible samples. On the contrary, our method satisfies these properties, thus creating more realistic samples.  
    \item We also show that both notions of coherence and association are positively correlated with predictive performances. Thus, creating implausible samples not only makes the models less trustworthy, but also reduces the performances.
    \item We compare our proposed method with other rebalancing strategies on two banking public data sets and one private data set from a major financial institution. We show that our proposed strategy, MGS-GRF, which is both coherent and preserves association, has the best predictive performances.  
\end{itemize}

%% file: related_work.tex
\section{Related work}
\label{sec:related_work}

\paragraph{Notations} 


We consider a data-set $\{(X^i,Y^i)\}_{i=1}^{N}$ constituted of $N$ independent pairs, each one distributed as $(X,Y)$. The random variable $X$ takes values in $\mathds{R}^{d} \times \mathcal{X}^p$ while $Y \in \{0,1\}$, where $\mathcal{X}$ is the space of categorical features. Here, without loss of generality, we assume that the first $d$ features of $X$, denoted $X_{1:d}$, are continuous, while the $p$ others, denoted $X_{d:}$, are categorical. 
Similarly, we suppose that the $n$ first samples are labeled $Y=1$, denoted $\{X^i\}_{i=1}^{n}$, verifying $n << N-n$ since we work in an imbalanced data setting.

\paragraph{\Cref{alg:mete_oversampling}} 
All rebalancing strategies in the literature are divided into two parts, the first one handling the continuous features and the second one handling the categorical ones. Accordingly, we encompass all oversampling strategies in \Cref{alg:mete_oversampling}, which describes the generation of a single synthetic example. \Cref{alg:mete_oversampling} may be run as many times as necessary to obtain the desired number of minority samples.
The procedure starts by selecting uniformly at random $c \in \{1, \hdots, n\}$ and the corresponding minority sample $X^c$.
Let $\textrm{NN}_{K, L}(X^c)$ be the set of the $K \in \mathbb{N}^*$ nearest neighbors of $X^c$ among minority samples w.r.t. to a given norm $L$. 
Finally, \texttt{ContinuousSampler} and \texttt{CategoricalSampler} functions are applied if necessary. We present below the state-of-the art procedures using \Cref{alg:mete_oversampling}.

SMOTE \cite{chawla2002smote} is the most common synthetic procedure for generating continuous new samples in the minority class. Thus, SMOTE does not have a \texttt{CategoricalSampler} in \Cref{alg:mete_oversampling}. To generate a new synthetic sample, an observation $X^k$ is drawn uniformly at random among the $K$ nearest neighbors of $ X^c$ w.r.t. the $L_2$ norm. The synthetic sample generation is the following 
\begin{align}
& \texttt{ContinuousSampler}\left(X^c_{1:d},\textrm{NN}_{K, L_2}(X^c)\right) = X^c_{1:d} + wX^{k}_{1:d}, \nonumber 
\end{align}
where  $ w \sim \mathcal{U}([0,1])$ and with $\mathcal{U}$ the uniform distribution. Note that SMOTE has several variants \cite{han2005borderline,he2008adasyn}, but, to the best of our knowledge, these variants are also originally designed for continuous input features only. 
\begin{algorithm}[t]
   \caption{OverSampler: One iteration for generating a new sample.}
   \label{alg:mete_oversampling}
\begin{algorithmic}
    \REQUIRE $X^1, \hdots, X^n$, \texttt{ContinuousSampler} and \texttt{CategoricalSampler}, $d$, $p$.
    
    \STATE Select uniformly $X^c$ among $X^1, \hdots, X^n$.
    \STATE Derive $\textrm{NN}_{K,L}(X^c)$ the set composed of the $K$ nearest-neighbors of $X^c$.
    \IF{$d > 0$ :} 
    \STATE  $Z_{1:d} \gets \texttt{ContinuousSampler}\left(X^c_{1:d},\textrm{NN}_{K,L}(X^c)\right)$.
    \ENDIF
    \IF{$p> 0$ :}
    \STATE  $Z_{d:} \gets \texttt{CategoricalSampler}\left(\textrm{NN}_{K,L}(X^c)\right)$.
    \ENDIF
    \RETURN $Z = [Z_{1:d},Z_{d:}]$, new minority class synthetic sample.
\end{algorithmic}
\end{algorithm}

SMOTE-N is presented in the original paper introducing SMOTE \cite{chawla2002smote}. This methodology is designed only for categorical input. SMOTE-N uses a version of the Value Difference Metric \cite{stanfill1986toward}, denoted $L_{\textrm{\tiny VDM}}$, as norm. More precisely, for two categorical vectors $u,v \in \mathcal{X}^p$ we have, 
\begin{align}
L_{\textrm{\tiny VDM}}(u_j,v_j) = \sum_{j=1}^{p} \delta(u_j,v_j),  \nonumber 
\end{align}
where 
$\delta(u_j,v_j) = 2 |p_n(Y=0|X_j=u_j) - p_n(Y=0|X_j = v_j)|. $ 
The value $p_n(Y=0|u_{j}) $ is the empirical conditional probability that the output class is $Y=0$ given that the feature $j$ has the value $u_{j}$. Note that, in order to compute $L_{\textrm{\tiny VDM}}$, the majority class samples are necessary. To generate a new observation, a sample is drawn uniformly among the minority samples. Then, its nearest neighbors according to $L_{\textrm{\tiny VDM}}$ are computed. Finally, the new minority sample is generated by a vote among the previous nearest neighbors along each variable.
With \Cref{alg:mete_oversampling} notations :
\begin{align}
& \texttt{CategoricalSampler}\left(\textrm{NN}_{K, L_{\textrm{\tiny VDM}}}(X^c)\right) = \textrm{Vote}\left(\textrm{NN}_{K, L_{\textrm{\tiny VDM}}}(X^c)\right),  \nonumber 
\end{align}
where $\textrm{Vote}\left(\textrm{NN}_{K, L_{\textrm{\tiny VDM}}}(X^c)\right)_j $, is a vote among the nearest-neighbors for the categorical feature $j \in \{1, \hdots, p\}$.

SMOTE-NC is designed to handle data sets containing both continuous and categorical features, and is also presented in the original SMOTE paper \cite{chawla2002smote}. The main idea is to define a distance metric, denoted $L_{\textrm{\tiny NC}}$, that takes into account the categorical features.
To this aim, the  median $C \in \mathbb{R}$ of standard deviations of all continuous features for the minority class is computed. The $L_{\textrm{\tiny NC}}$ takes the form
\begin{align}
L_{\textrm{\tiny NC}}(X,X^{\prime}) = \sqrt{ \sum_{j=1}^{d} \left(X^{\prime}_j -X_j\right)^2  +  C^2\sum_{j=d+1}^{d+p} \mathds{1}_{X_j \neq X^{\prime}_j} }.    \nonumber
\end{align}
Then, continuous features are generated using SMOTE interpolation while categorical one are based on a nearest neighbors vote.
For SMOTE-NC we have: 
\begin{align}
& \texttt{ContinuousSampler}\left(X^c_{1:d},\textrm{NN}_{K, L_{\textrm{\tiny NC}}}(X^c)\right) = X^c_{1:d} + wX^{k_{c}}_{1:d}, \nonumber \\
& \texttt{CategoricalSampler}\left(\textrm{NN}_{K, L_{\textrm{\tiny NC}}}(X^c)\right) = \textrm{Vote}\left(\textrm{NN}_{K, L_{\textrm{\tiny NC}}}(X^c)\right), \nonumber
\end{align}

SMOTE-ENC \cite{mukherjee2021smote} applies the same procedure as SMOTE-NC except that $L_{\textrm{\tiny NC}}$ is replaced by 
\begin{align}
L_{\textrm{\tiny ENC}}(X,X^{\prime}) = \sqrt{ \sum_{j=1}^{d} \left(X^{\prime}_j -X_j\right)^2  +  \sum_{j=d+1}^{d+p} C_j^2\mathds{1}_{X_j \neq X^{\prime}_j} }.    \nonumber
\end{align}
However, to the best of our knowledge, SMOTE-ENC has no implementation agnostic to the data-set, and in the original repository the computation of $C_j$ differs for each data-set.

%% file: algorithm.tex
\section{Our proposed algorithm: MGS-GRF}
 
In this section, we describe our new algorithm to handle mixed data. It is organized similarly to \Cref{alg:mete_oversampling}, so that we  first describe  our procedure to generate continuous features before detailing the categorical variables methodology.

\subsection{Handling continuous features}
\label{sec:alg-cont-features}

Numerous studies have proposed Kernel Density Estimator (KDE) for generating synthetic samples within the minority class for continuous input features. Actually, SMOTE itself can be seen as a KDE with uniform kernel piece by piece \cite{stocksiekergeneralized}.
For instance, \cite{lee1999regularization,lee2000noisy} introduce an oversampling strategy that, based on original samples, adds centered Gaussian noise with a unique diagonal scale matrix to original samples to generate new observations.
\cite{menardi2014training} develops ROSE, a KDE based oversampling strategy which is associated to a unique scale matrix for the whole minority class.
Later, \cite{tang2015kerneladasyn} proposes a weighted sample KDE oversampling strategy with fixed diagonal scale matrix, thus isotropic, of the form $h \times I$ with $h\in \mathds{R}$ and $I$ the identity matrix.
$h$ is as in Adasyn \cite{he2008adasyn}: higher weights are given to original minority samples surrounded mostly by majority class samples. 


Multivariate Gaussian SMOTE (MGS) is a synthetic procedure for continuous features introduced by \cite{sakhowe}. MGS is presented as a variant of SMOTE that generate new samples from multivariate Gaussian distributions and no longer with a linear interpolation. We analyzed the MGS procedure and reformulate it as a Gaussian KDE from the sample smoothing estimator family \cite{scott2015multivariate}, i.e. with several different full-rank local scale matrices. Furthermore, MGS do not assume the covariance matrix to be diagonal, thus not isotropic, which allows for better adaptivity to the unknown minority distribution.
We choose to generate the continuous features of synthetic samples, $Z_{1:d} \in \mathds{R}^d$, according to the following density $\hat{f}_{MGS}(Z_{1:d})$ fitted on the original minority samples $\{X^i\}_{i=1,\hdots,n}$:
\begin{align}
    \hat{f}_{MGS}(Z_{1:d}) &=\frac{1}{n}\sum_{i=1}^{n}\frac{1}{(2\pi)^{d/2}|{\hat{\Sigma}^i}|}\exp\left(-\frac{1}{2}(Z_{1:d}-\hat{\mu}^i)^{T}(\hat{\Sigma}^i)^{-1}(Z_{1:d}-\hat{\mu}^i)\right),
    \label{eq_def_mgs}
\end{align}
where $|\hat{\Sigma}^i|$ denotes the determinant of $\hat{\Sigma}^i$ and
\begin{align*}
\hat{\mu}^i = \frac{1}{K} \sum\limits_{X  \in \textrm{NN}_{K, L_2}(X^{i})} X_{1:d}, \quad   \hat{\Sigma}^i = \frac{1}{K} \sum\limits_{\substack{X  \in \textrm{NN}_{K, L_2}(X^{i})}} \left(X_{1:d} - \hat{\mu}^i\right)\left(X_{1:d} - \hat{\mu}^i\right)^T   
\end{align*} 
are estimated for each minority class sample using the $K$ nearest-neighbors from the minority class, w.r.t. to $L_2$ norm.
We choose a value of $K=d+1$ in order to possibly obtain full-rank covariance matrices. Besides, $\Sigma^i$ can be estimated using shrinkage 
\cite{ledoit2004well,chen2010shrinkage} or simply the sample covariance matrix \cite{sakhowe}, but empirically we got better results  with the empirical  covariance.

One notes that the underlying distribution of this sampling is a $n$ Gaussian mixture with equal weights.
We also remark that ROSE \cite{menardi2014training} corresponds to the special case where all $\Sigma^i$ are equal.


\subsection{Handling categorical data via Generalized Forests}
\label{sec:categorical}

Now, we introduce our selected procedure for generating synthetic categorical features.
A first remark when looking at  \Cref{alg:mete_oversampling} is that multi-output classifiers, such as nearest neighbors, can be used to generate the categorical features. Indeed, such models can be trained using only minority samples, aiming at predicting the categorical features  $\{X^i_{d:}\}_{i=1}^{n}$ based on the continuous features $\{X^i_{1:d}\}_{i=1}^{n}$. 
Denoting by $\hat{g}$ such trained classifier, based on ~\Cref{alg:mete_oversampling}, one can repeatedly generate categorical samples as 
\begin{align*}
    \texttt{CategoricalSampler}( Z_{1:d}) = \hat{g}(Z_{1:d}).
\end{align*}
Our selected methodology to generate categorical variables relies on Generalized Random Forests (GRF) \cite{athey2019generalized}.
The main difference between a random forest \cite{breiman2001random} and a GRF, is that, given the new point, GRF assigns a probability to each training sample. 
These probabilities are derive from the frequency of the training samples to fall in the same leaf as the predicted sample. Finally, GRF can be used to estimate any quantity identified via local moment conditions. 
\begin{algorithm}[t]
   \caption{Prediction procedure of GRF}
   \label{alg1:dRF}
    \begin{algorithmic}
       \REQUIRE Forest composed of $T$ trees $\mathcal{T}_1,\hdots,\mathcal{T}_T$. A new unlabeled sample $Z_{1:d}$.
       \STATE $\forall k=1..T, \,\,\mathcal{L}_k(Z_{1:d}) \gets$ set of training samples which end up in the same leaf as $Z_{1:d}$ in the tree $\mathcal{T}_k$
       \FOR{$i \in [1,\hdots,n]$}
        \STATE $w_{(Z_{1:d})}(X^i) \gets \frac{1}{T} \sum_{k=1}^{T} \frac{\mathds{1}_{\{X^i \in \mathcal{L}_k(Z_{1:d})\}}}{|\mathcal{L}_k(Z_{1:d})|}$
       \ENDFOR
       \STATE $Z_{d:}$ $\gets$ Sample $\{X^1_{d:},\hdots,X^n_{d:}\}$ based on $\{w_{(Z_{1:d})}(X^1_{1:d}), \hdots, w_{(Z_{1:d})}(X^n_{1:d})\}$.
       \RETURN $Z_{d:}$
    \end{algorithmic}
\end{algorithm}

We implemented our own version of GRF from the \textit{RandomForestClassifier} class of scikit-learn \cite{scikit-learn}. In our algorithm, the derivate probabilities are used to draw predicted target from training target vectors ($Y^i$). The predict procedure of our GRF is detailed in \Cref{alg1:dRF}. 
Besides, we try several default hyperparameters for our GRF and finally we keep the default values from \textit{RandomForestClassifier} class of scikit-learn for the tree building. Furthermore, we do not apply the principle of honesty \cite{biau2012analysis}, and neither scale the target variables.


\subsection{MGS-GRF}

We now detail MGS-DRF, our new procedure that combines MGS and GRF as described above.
It follows the three following steps.
First, MGS is applied to generate the continuous features of the new synthetic samples. 
Then, a Generalized Random Forest (GRF) denoted by $\hat{g}_{GRF}$ is trained on all the original minority samples with the continuous features $\{X^i_{1:d}\}_{i=1}^{n}$ as inputs and the categorical features $\{X^i_{d:}\}_{i=1}^{n},$ as outputs. Finally, the trained GRF is used to build the categorical features based on the continuous ones generated in the first step. Using ~\Cref{alg:mete_oversampling} notations we have,
\begin{align}
    &  \texttt{ContinuousSampler}(\{X^i_{1:d}\}_{i=1}^{n}) =Z_{1:d}  \sim \hat{f}_{ \textrm{\tiny MGS}} \nonumber \\
    &\texttt{CategoricalSampler}( Z_{d:}) = \hat{g}_{\textrm{\tiny GRF}}\left(Z_{1:d}\right). \nonumber
\end{align}

Our proposed method enjoys the following properties: $(i)$ GRF generates combinations of categorical features that are all from the original minority class. $(ii)$ Due to tree building procedure, GRF may be able to use only the few continuous variables that are relevant to generate the categorical variables, thus ensuring a better correlation between continuous and categorical variables. $(iii)$ The categorical features are generated directly from the continuous ones of the new sample. Thus, they are no longer based on the neighborhood of the central point.

%% file: illustrations.tex
\section{Illustrations on simulated data}

In the following, we describe our baselines before defining both coherence and associations. We illustrate these notions through numerical simulations. \footnote{All our experiments are available at \url{https://github.com/artefactory/mgs-grf}.}
%

\subsection{Baselines}

Now, we introduce different strategies to preprocess the original imbalanced data set. 
We denote by None strategy the procedure where no rebalancing strategy is applied. 
CW is the class-weighting strategy while Random Oversampling strategy (ROS) and Random Undersampling Strategy (RUS) are  
data-level approaches. 
We also include the synthetic procedure SMOTE-NC, with the default number of nearest neighbors equal to $5$. There is no generic implementation of SMOTE-ENC, thus we do not include this strategy (see \Cref{sec:related_work}).


Besides, we introduce $3$ synthetic baselines for our comparison. 
MGS-NC selects a central point $X^c$ uniformly over minority samples. MGS distribution is used (see Equation~\ref{eq_def_mgs}) with $\hat{\Sigma}^i$ and $\hat{\mu}^i$ computed on the $K$ nearest neighbors $NN_{L_{\textrm{\tiny NC},K}}(X^c)$ of $X^c$ w.r.t. the $L_{\textrm{\tiny NC}}$ norm. Then, each categorical variable is generated separately via a vote among the same neighbors.
The second baseline, MGS-5NN, applies MGS on the continuous features, and builds the categorical ones using a $k=5$ nearest neighbors w.r.t. $L_2$ norm as multi-output classifier $\hat{g}$ (see Section \ref{sec:categorical}). Similarly, MGS-1NN is the same procedure with $k=1$.

All strategies (except None) resample or generate observations so that each of the two classes contains the same number of observations (balanced data set). 


\subsection{Numerical illustrations of non-coherence notion}
\label{subsec:sim2}

In our first experimental protocol, we want to analyze the distribution of categorical variables via the notion of coherence defined below. 
 
\begin{definition}
We denote by $\mathcal{C}$ the set of combinations of categorical features in the original data set. We denote by $\mathcal{C}_{Y=1}$ the combination present in the original minority class $Y=1$. 
We say that a synthetic oversampling strategy is coherent, with respect to the minority class, if all combinations of generated categorical features belong to $\mathcal{C}_{Y=1}$. Accordingly, we say that a minority sample is coherent if its categorical vector belongs to $\mathcal{C}_{Y=1}$. 
\end{definition}

Our main objectives are to detect non-coherent synthetic procedures and assess whether incoherent samples harm predictive performances.
We define the coherence value, denoted $Coh$, by the proportion of coherent synthetic observations generated by a strategy, over all the synthetic data. If we denote by $n_{g}$ the number of generated samples $Z_{d:}^{\ell}$, we have $$Coh = \frac{1}{n_{g}} \sum_{\ell=1}^{n_{g}} \mathds{1}_{\{Z_{d:}^{\ell} \in \mathcal{C}_{Y=1} \}}. $$

We note that strategies that generate categorical features one by one with a vote are not coherent, as they can mix original combinations. This applies to SMOTE-NC, MGS-NC and MGS-5NN. However, MGS-1NN copies the features of the nearest neighbor from the minority class, thus leading the combination to be originally present in the minority class. Similarly, GRF is coherent because it draws randomly a combination of categorical features from the minority class.

\paragraph{Protocol}

We will simulate a binary classification task data set such that class $0$ is overrepresented, with $d=9$ continuous features and $p=2$ categorical ones.
We denote by $\mathcal{C} = \mathcal{D} \times \mathcal{E}$ the set of combinations of categorical features where $\mathcal{D}$ (resp. $\mathcal{E}$) is the set of possible modalities for the first (resp. second) categorical features. Each categorical feature is composed of $m$ modalities, i.e. $|\mathcal{D}| = |\mathcal{E}| = m$ and $|\mathcal{C}|= m^2$, and only $m$ (out of $m^2$) combinations of categorical features are present in the minority class, written $\mathcal{C}_{Y=1}=m$.
Only the $3$ informative continuous features and the categorical features are used for generating the target $Y$.
Our procedure consists of the following steps :
\begin{enumerate}
    \item Draw $5000$ samples composed of $d$ continuous features as follows $X_{1:d} = ( X_1,\hdots,X_{d}) \sim \mathcal{N}(0, I_{d})$.
    \item Draw $Z \in \mathcal{C}$ such that 
        \begin{align}
        \mathds{P}[Z = c | X_{1:d}] = \frac{\exp(-\theta_c^\top X_{1:3})}{  \sum_{\ell \in \mathcal{C}} \exp(-\theta_{\ell}^\top X_{1:3})}, \nonumber
        \end{align}
    with c $\in \mathcal{C}$. The set of parameters $\Theta =\{ \theta_c, c \in \mathcal{C}\}$, verify, for all $c \in \mathcal{C}$, $\theta_{c} \in \mathds{R}^{3}$. $X_{1:3}$ are the 3 informative components of $X$, while other $X_{j}$ values ($j>3$) do not impact $Z$ value.
    \item Draw the target variable $Y$ such that $$Y|X_{1:d},Z\!=\!c \, \sim \, \mathcal{B}(\sigma(\alpha^\top X_{1:3}+ \gamma_{c})),$$ where $\mathcal{B}$ is the Bernoulli distribution, $\sigma$ is the logistic function and $\alpha \in \mathds{R}^3$. The set of parameters $\Gamma = \{ \gamma_c, c \in \mathcal{C}\}$ verify, for all $c \in \mathcal{C}$, $\gamma_{c} \in \mathds{R}$. In order to limit the number of coherent combinations present in the minority class, we set high $\gamma_c$ values only for $m$ different combinations $c \in \mathcal{C}$. Besides, we choose $\alpha$ and all $\gamma_c$ values such that the class $Y=1$ is underrepresented. 
    \item Return $[X_{1},\hdots,X_{d},Z_1,Z_2,Y]$, where $Z_1 \in \mathcal{D}, Z_2 \in \mathcal{E}$ satisfy $Z=(Z_1, Z_2)$.
\end{enumerate}
 

\begin{table}[b]
\centering
\caption{LightGBM trained on simulated data from experimental protocol. Standard deviations are available in \Cref{tab:sim2-3-annexe} in \Cref{app:supp-mat}.
}
\label{tab:sim2-3-corps}
\setlength\tabcolsep{2.5pt}
\begin{tabular}{lccccccccc}
     \toprule 
       Strategy &  None    &  CW     &  ROS & RUS &    SMOTE  &MGS&   MGS &MGS   &    MGS \\ 
       &         &  & & & -NC  &  -NC & -5NN &-1NN   &-GRF \\ 
     \midrule
     \small PR AUC &	0.903	&0.903	&0.893	&0.699	&0.860	&0.922	&0.870	&0.952 &\textbf{0.954} \\
     \small ROC AUC   &0.975	&0.977	&0.975	&0.935	&0.962	&0.984	&0.970	&\textbf{0.993}	&\textbf{0.993}\\
     Coh & \textbf{100}\%	& \textbf{100}\% & \textbf{100}\% &\textbf{100}\%	& 90\%	& 90\%	&	83\% &	\textbf{100}\% &\textbf{100}\%	\\ 
     Time (s) &0.55	&0.56	&0.74	&0.27	&1.01  &1.23  &1.04   &1.00   &{1.43} \\
     \bottomrule
\end{tabular}
\end{table}
\paragraph{One combination of $\Theta,  \alpha, \Gamma$}
We fix the values of $\Theta$, $\alpha$ and $\Gamma$ once and for all and run the above protocol $50$ times. Thus, we produce $50$ data sets (with different random seeds), that we split into train and test set. We preprocess the train set with the different rebalancing strategies and train a LightGBM classifier on it. Predictive performances on the test set are displayed in \Cref{tab:sim2-3-corps}.

In \Cref{tab:sim2-3-corps}, we see that MGS-GRF and MGS-1NN have a coherence value $Coh=100\%$, which is expected for these two coherent strategies. On the contrary, the non-coherent strategies (SMOTE-NC, MGS-NC and MGS-5NN) have coherence values lower than $100\%$, as they can generate minority samples whose categorical vectors are not found in the original data set. 
We observe that strategies with low $Coh$, i.e. creating non-coherent combinations of categorical features deteriorate the predictive performances of the final classifier. 
This is particularly visible for MGS-5NN and MGS-1NN, while being very similar models.
In contrast, MGS-GRF achieves the best predictive performances in terms of both PR AUC and ROC AUC, with a computation time (for oversampling and LightGBM training) only $50\%$ longer than SMOTE-NC. 
Finally, we remark that MGS-NC leads to better predictive performances than SMOTE-NC, indicating that MGS seems to better regenerate the distributions of the minority class than SMOTE.

\begin{figure}[t]
\centering
    \begin{subfigure}[b]{0.48\textwidth}
        \centering
        \includegraphics[width=0.98\textwidth]{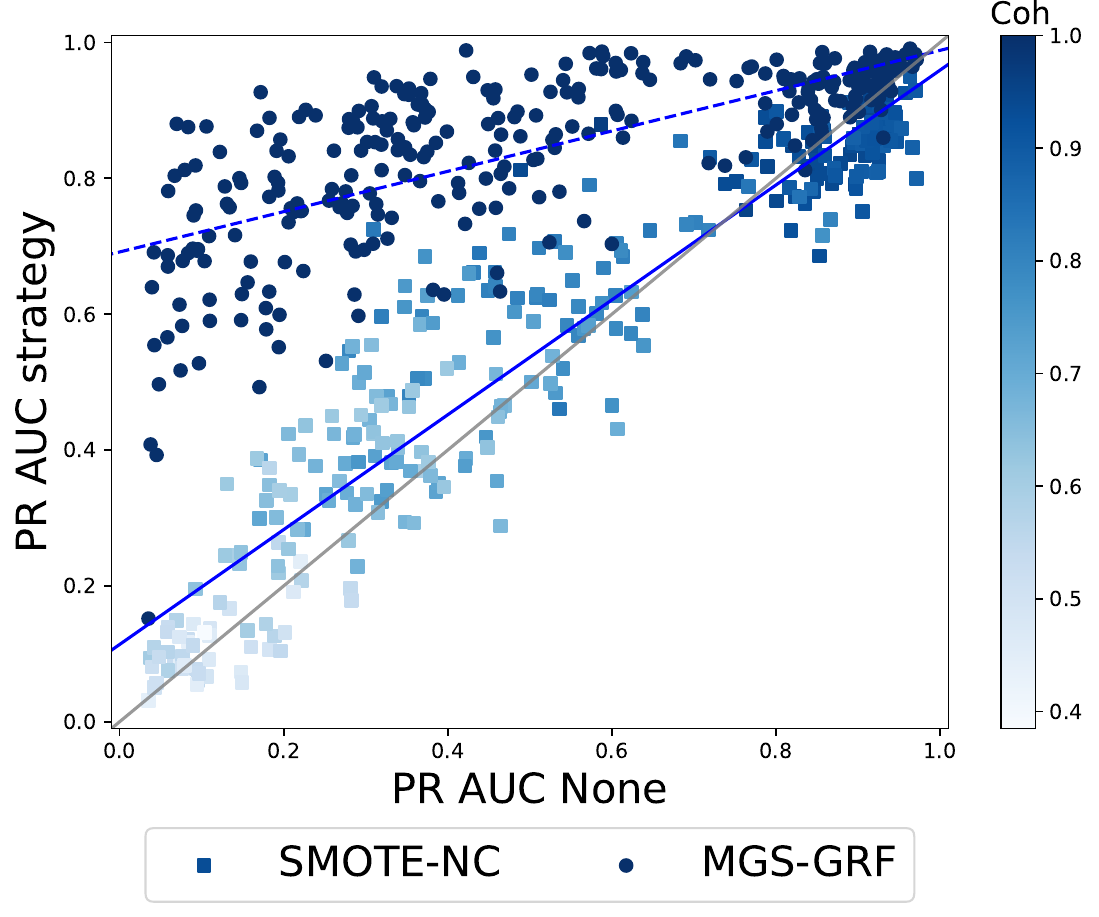}
        \caption{SMOTE-NC and MGS-GRF.}
        \label{fig:sim-2}
    \end{subfigure}
    \begin{subfigure}[b]{0.48\textwidth}
        \centering
        \includegraphics[width=0.98\textwidth]{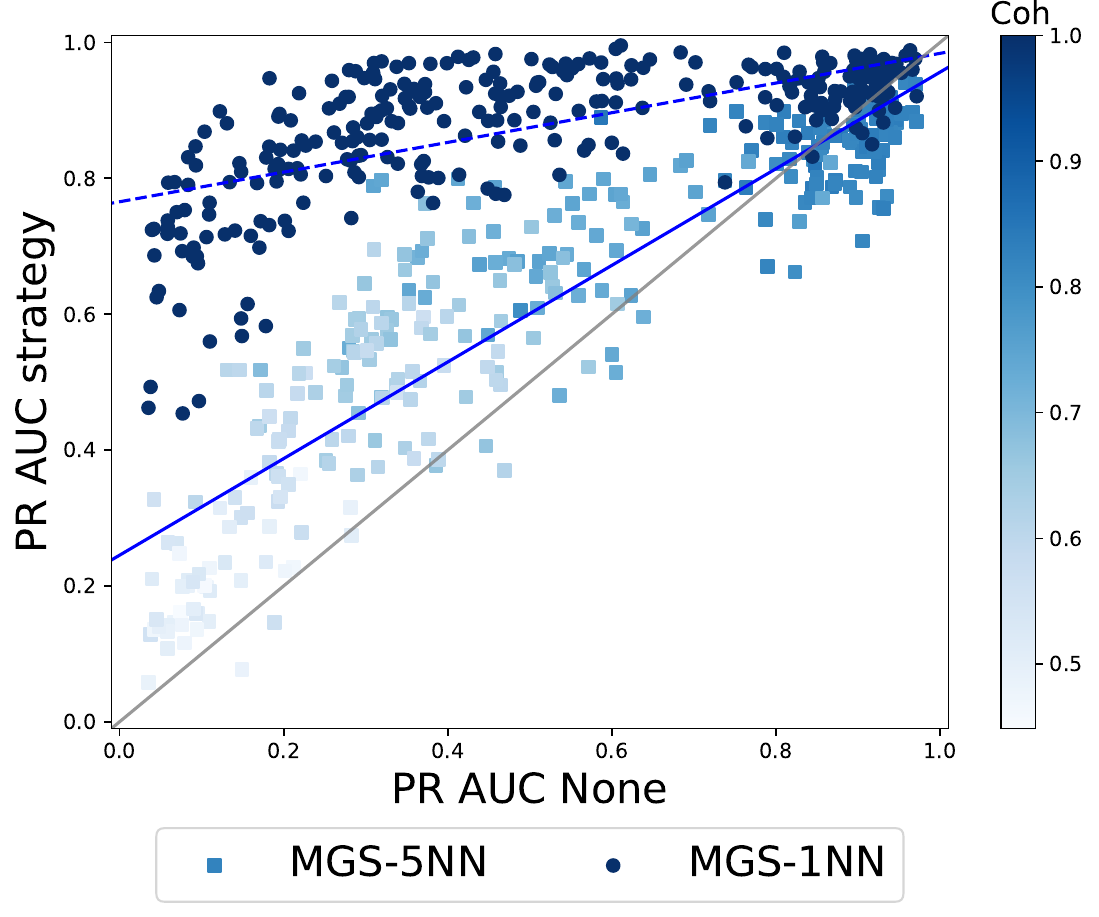}
        \caption{MGS-5NN and MGS-1NN.}
        \label{fig:sim-2-2}
    \end{subfigure}
\label{fig:sim-coherence}
\caption{$PR \; AUC$ of coherence simulations. Points color reflect their $Coh$ value.}
\end{figure}
\paragraph{Different combination of $\Theta,\alpha$ and $\Gamma$}
We run our protocol with 6 configuration values for $\Theta,\alpha,\Gamma$. For each configuration, we apply the protocol above, so that we obtain in total $300$ datasets. The PR AUC of the LightGBM classifier for each rebalancing strategy and for each data set is displayed in \Cref{fig:sim-2,fig:sim-2-2}, where each point corresponds to one of the $300$ data sets. 
%
 We display the PR AUC of a given rebalancing strategy in y-axis and the PR AUC of the None strategy in x-axis.
Circles points are all associated to coherent strategies (MGS-1NN and our proposed strategy MGS-GRF), while the squares ones are associated to non-coherent ones (SMOTE-NC, MGS-5NN).
We plot linear fitting curves and also add the first bisector in gray (line $y=x$).


%

%
In both figures, we remark that the points (both squares and circles) are above the first bisector, thus the rebalancing strategies lead to improvement of PR AUC.
However, the average coherent strategies achieve higher PR AUC than the non-coherent ones. This difference is the highest when the PR AUC of the None strategy is the lowest (left side of the figures), which corresponds to more complex classification settings. 
In such difficult scenarios, non-coherent strategies have low $Coh$ values, which may in turn explain their low PR AUC, close to that of the None strategy. 
When the learning task is easier, all strategies  have similar performance (right side of the figures). All in all, this experiment shows that coherent strategies should be preferred to non-coherent ones, especially in more difficult classification problems.





\subsection{Numerical illustrations of association notion}
\label{subsec:sim-high-dim}

In the following, we define and present our numerical experiments on association.


\begin{definition}
    The association level of a multi-output classifier is its predictive performance when inferring categorical features with respect to continuous ones. This performance is measured as the empirical excess risk w.r.t. Bayes error.
\end{definition}
We choose to measure the association level of a classifier via its accuracy on a leave-one-out validation on original minority samples. 
%
More precisely, if we write $\hat Z^{\ell}_{d:}$ the prediction for $X_{1:d}^\ell$ of a given classifier trained on $\{X^i\}_{i\neq \ell}$ (leave-one-out prediction), the association level $Asso$ of this classifier is 
\begin{align}
    Asso = 1 - \left(\frac{1}{n} \sum_{\ell=1}^{n} \mathds{1}_{\{X^{\ell}_{d:} \neq \hat Z^{\ell}_{d:}\}}
    - \frac{1}{n} \sum_{\ell=1}^{n} \mathds{1}_{\{X^{\ell}_{d:} \neq h^*(X^i_{1:d})\}}\right),  \nonumber
\end{align}
where  $h^*$ is the Bayes classifier whose predictions are defined by $h^*(X_{1:d}) = \arg \max_{c \in \mathcal{C}}\mathbb{P}(X_{d:}=c|X_{1:d})$, with $\mathcal{C}$ the set of combination of categorical features.
In practice, we do not have access to the Bayes classifier, and thus to the association level. In such situations, the association level can also be estimated without the last term, that is with the classifier accuracy. Another point is that we focus on original minority points for which we have a ground truth $X^{i}_{d:}$. While measuring on generated continuous features would be ideal given our oversampling objective (see Section \ref{sec:alg-cont-features}), we do not have the ground truth for the categorical values of those points, and all reference data for measuring association are with the minority points.

\begin{figure}[t]
\centering
    \begin{subfigure}[b]{0.48\textwidth}
        \centering
        \includegraphics[width=0.91\textwidth]{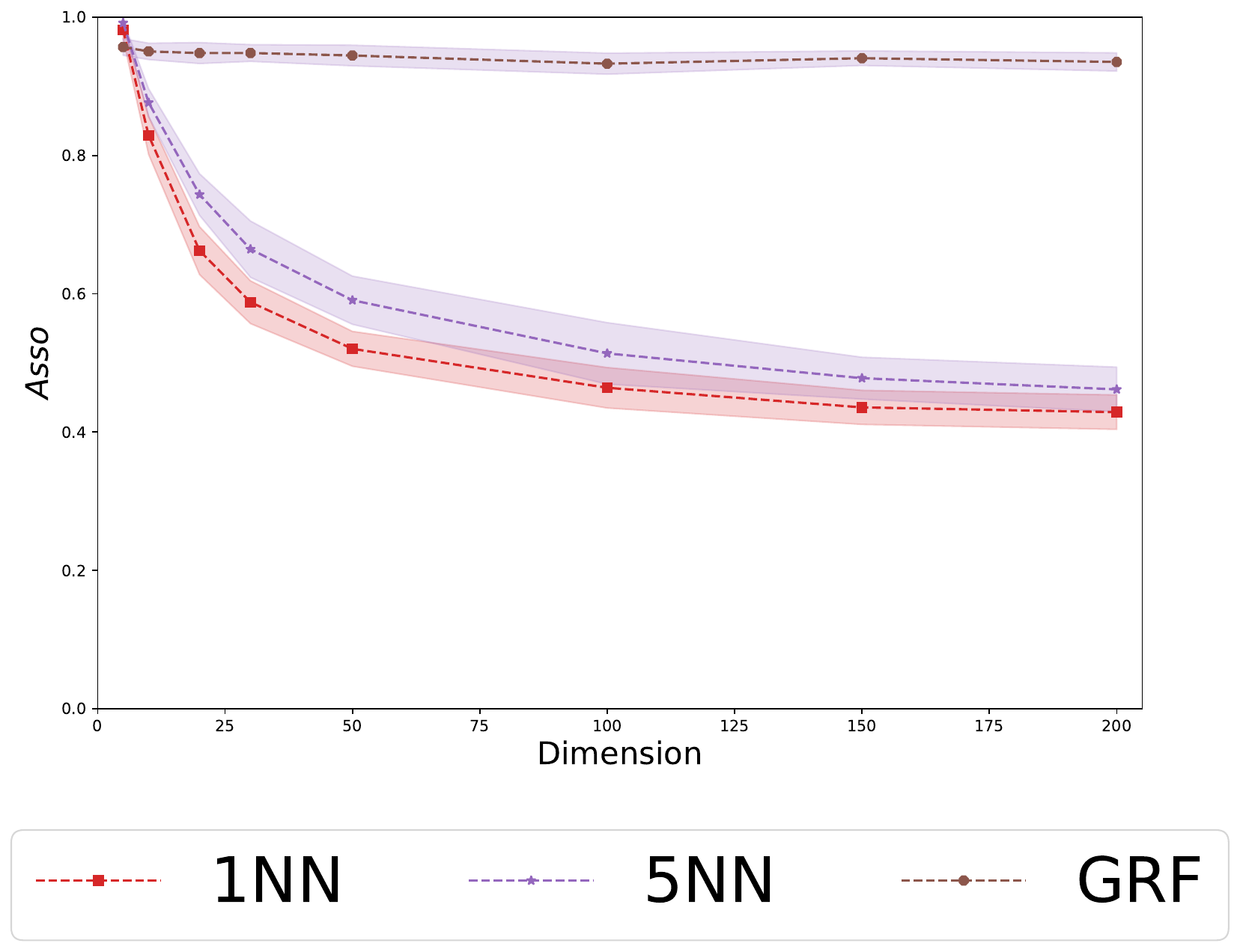}
        \caption{$Asso$ w.r.t. to dimension.}
        \label{fig:sim-high-dim-asso}
    \end{subfigure}
    \begin{subfigure}[b]{0.48\textwidth}
        \centering        \includegraphics[width=0.91\textwidth]{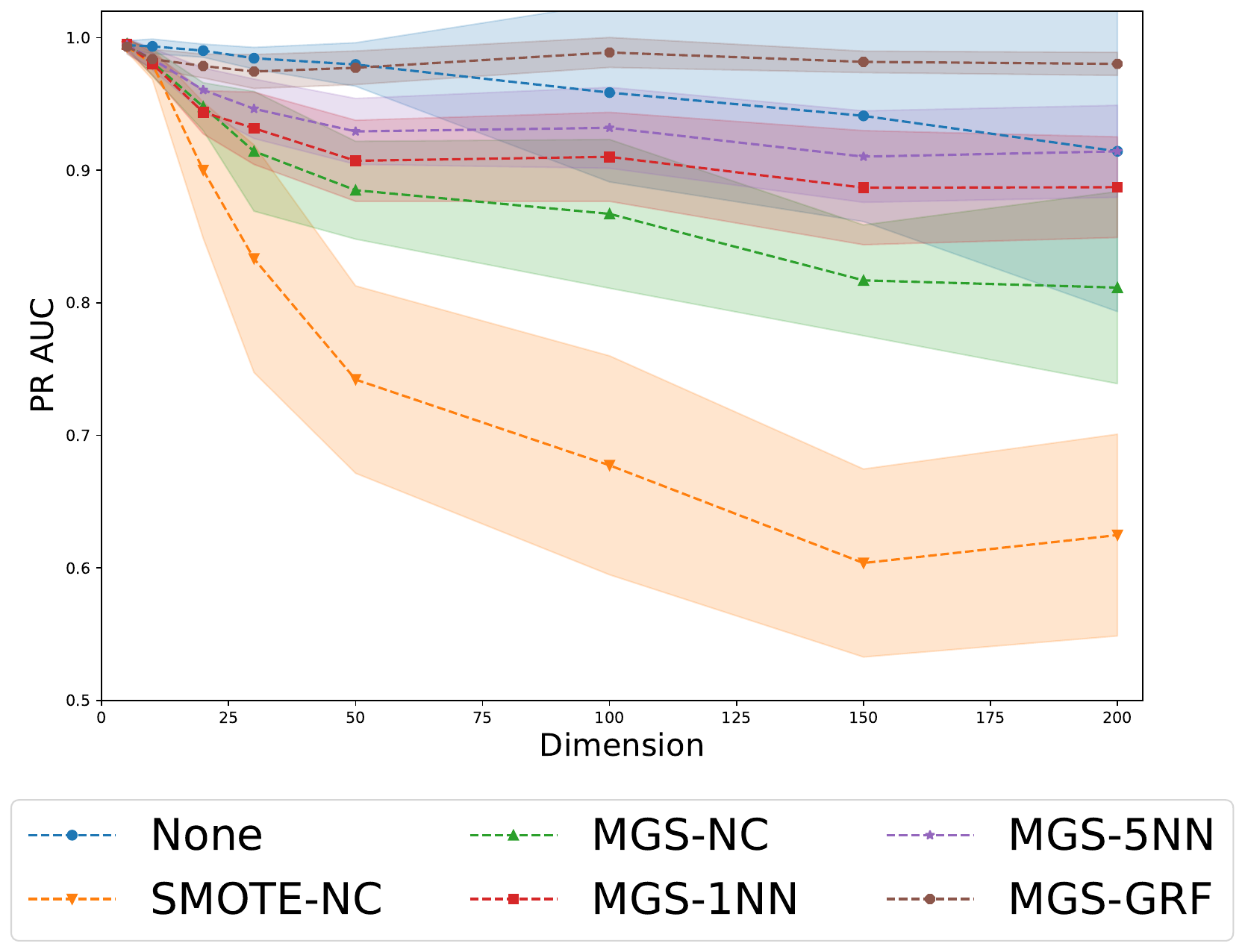}
        \caption{PR AUC w.r.t. to dimension.}
        \label{fig:sim-high-dim}
    \end{subfigure}
\label{fig:sim-assocation-exp}
\caption{Association experiments in high dimensional setting with noisy features.}
\end{figure}

In this second numerical experiments, we generate an imbalance binary classification data based on four input variables: 3 of them are continuous and the remaining one is categorical, with $3$ modalities.  
We add $d-3$ continuous noise variables, which are independent of all previous variables. More precisely:
\begin{enumerate}
    \item Draw $5000$ samples as a mixture of $3$ Gaussian in $\mathds{R}^3$: $(X_1,X_2,X_3) \sim \sum_{w=1}^3 \pi_w \mathcal{N}(\mu_w, \Sigma_w)$, with $\sum_{w=1}^3 \pi_w = 1$ and $\pi_w \geq 0$. Let $W \in \{1,2,3\}$ be the latent variable of the mixture s.t. $(X_1,X_2,X_3)| W \sim \mathcal{N}(\mu_W, \Sigma_W)$. 
    \item Draw $d-3$ noise features: 
    $(X_4,\hdots,X_{d}) \sim \mathcal{N}(\mu_2, \lambda I_{d-3})$ with $\lambda \in \mathbb{R}^*$.
    \item Draw $Z \in \{``A",``B",``C"\}$ such that,
    \begin{align}
        \mathds{P}[Z = c | X_{1:d},W=w] = \frac{\exp(-\zeta_{c}^\top X_{1:3} + \chi_{w,c})}{  \sum_{\ell \in \mathcal{C}} \exp(-\zeta_{\ell}^\top X_{1:3} + \chi_{w,\ell})}, \nonumber
    \end{align}
    where $\zeta_{c} \in \mathds{R}^3$ and $\chi_{w,c} \in \mathds{R}$.  For each Gaussian, that is for each $w \in \{1,2,3\}$, we choose $\chi_{w,c}$ such that one modality is associated to the minority class. We emphasize that the notion of \textit{association} between categorical and continuous features occurs at this step, where $Z$ depends only on the 3 informative values $X_{1:3}$ ($W$ is a confounding variable) while others $X_{j} (j>3)$ are pure noise.
    \item Draw the target variable $Y$ such that $$Y|X_{1:d},W\!=\!w,Z\!=\!c \, \sim \, \mathcal{B}(\sigma( \beta^\top X_{1:3} + \eta_{w} + \phi_{c})),$$ where $\beta \in \mathds{R}^3$, and $ \eta_w, \phi_c  \in \mathds{R}$. One notes that $Y$ depends on $X_{1:3}$, while other $X_j$ ($j>3$) are  non-informative.
    
    \item Return $[X_1,X_2,X_3,X_4, \hdots, X_{d},Z,Y]$.
\end{enumerate}
Following this protocol, we generate $8$ data sets with increasing number of non-informative features $d-3$. 
On each of these $8$ data sets, we estimate the association level of $3$ multi-output classifiers: 1NN, 5NN and GRF. Since we are in a simulation setting, we have access to the Bayes predictor (see \Cref{sec:app-sim-high-dim} for details), which allows us to compute the association level. 
We observe in \Cref{fig:sim-high-dim-asso}, that the association level of nearest neighbors (1NN and 5NN) decreases with increasing dimension, contrary to that of GRF which remains constant, and close to $1$, this more accurately generating categorial features. As it is a known behavior for supervised tasks with many noisy uninformative features, nearest neighbors do not predict well the categorical feature.

We now study how the initial prediction task (predicting $Y \in \{0,1\}$ based on continuous and categorical features) is impacted by the categorical feature generation. 
On each data set, we apply rebalancing strategies followed by LightGBM (with default hyperparameters) and compute its PR AUC. Results are depicted in \Cref{fig:sim-high-dim}.
We observe that all methods have the same performance for low dimensions. In this setting, the problem can be considered as easy (since the None strategy has good performances) and all rebalancing strategies are roughly equivalent, similarly to experiments implemented in Section~\ref{subsec:sim2}. 

As expected due to the curse of dimensionality, all performances degrade when the dimension increases, with the notable exception of our proposed method MGS-GRF, whose performances remain unaffected by the addition of noise variables. In fact, we see that the use of GRF compared to nearest neighbors (MGS-1NN or MGS-5NN) for generating categorical variables improves the final predictive performances. This finds explanation in the splitting procedure at work in GRF, which selects the variables that are the most predictive of the output (here the categorical input vector). On the contrary, nearest neighbors are unable to detect relevant variables for splitting, which explains their poor performances in high-dimensional settings. 

We also note that SMOTE-NC, probably the default synthetic rebalancing strategy, is the worst in high dimensions, both in terms of mean value and standard deviation. On the opposite, our proposed method MGS-GRF exhibits the best performances with a small standard deviation. This seems to indicate that a good generation of categorical features (via GRF) leads to good predictive performance on the initial binary classification task.


%% file: experiments.tex
\section{Experiments on real-world data sets}
\label{sec:exp-data-real}

\begin{wraptable}[9]{r}{0.51\textwidth}
    \centering
    \caption{Data sets.}
    \begin{tabular}{ccccc}
         \toprule
          &  $N$ & $n/N$ & $d$ & Cat   \\ 
         \midrule
         Private & $\simeq 10^7$ & $\,<\!1\%$ & $\,>\!200$ & $\,<\!10$ \\
         BankMarketing & 40325 & 1\%  & 16 & 10 \\
         BankChurners & 8585 & 1\%  & 19 & 5\\
        \bottomrule
    \end{tabular}
    \label{table:data-sets}
\end{wraptable}
In this section, we describe all our numerical experiments on real-world data sets. We describe our protocol before commenting our results.

\subsection{Data sets}


We use two open source banking-related data sets, Bankmarketing \cite{mukherjee2021smote} and Bankchurners \cite{ZhyliBankchurners}, described in \Cref{table:data-sets}, both about bank customer behavior prediction. The first data set objective is to predict if a client subscribes to a banking offer after a phone marketing campaign. The second data set aims at predicting customer attrition from a financial institution. Both data set covariates contain historical records of the customers.
To be closer to the challenge encountered in the private sector, we undersample the open source data set to have an imbalance ratio of 1\%. 

We also have a private data set, from a major bank, that contains clients information from one country in Europe. The purpose is to predict if a customer meets some criterion from historical records. The target criterion is beyond the scope of this paper. Positive cases predicted by the model are pushed to analysts, with corresponding explainability results, and the analysts have to make a decision based on the model output. Furthermore, analysts give feedbacks on the pertinence of pushed cases to the data science team, who retrain the model several times per year. A version from the ML-based system has been deployed and the high-level pipeline is described in \Cref{fig:pipeline} in \Cref{app:supp-mat}.  
The data set contains millions of \textit{anonymized} customers and we recall that all process is done in compliance with the country's regulatory requirements.


\subsection{Evaluation}

We evaluate the public data sets with the following protocol. For an iteration, the data set is evaluated through a $5$-fold cross validation, with Z-score scaling of the train set.  
We stress on the fact that each strategy is applied on the same training set.
We run this protocol $20$ times and averaged the metrics from each run. The private data set is evaluated through a temporal train/test split, with test set covering year 2023 and no overlap of clients between train and test sets.
Tree-based models produce state-of-the-art performances on tabular data sets \cite{grinsztajn2022tree} and we choose LightGBM \cite{ke2017lightgbm} as classifier due to its computational efficiency. 

We introduce an evaluation metric, the precision at recall, denoted  $\textrm{Pr-at-rec}(x)$, which equals the precision associated to a recall of at least $x$, for any $x \in [0,1]$. This metric aims at representing an industrial or operational trade-off between precision and recall. After discussions with the analyst, we choose a recall $x=0.2$.
We also use two usual aggregated metrics, the ROC AUC and the PR AUC. Results are displayed in \Cref{tab:merged_results}.

\begin{table}[t]
\centering
\caption{BankChurners, BankMarketing and Private data sets. For confidentiality motivations, private data set metrics are relative gains compared to None strategy and no running time is provided. Standard deviations are available in \Cref{tab:merged_results_std}.
}
\label{tab:merged_results}
\setlength\tabcolsep{2pt}
\begin{tabular}{llllllllllll}
     \toprule 
      \multirow{2}{*}{\footnotesize Metric} & \multirow{2}{*}{\footnotesize Data} & \multicolumn{9}{c}{Strategy} \\
      \cmidrule(lr){3-12}
       &  &  \footnotesize None    &  \footnotesize CW     &  \footnotesize ROS & \footnotesize RUS &   \scriptsize SMOT  & \footnotesize MGS &   \footnotesize MGS & \footnotesize MGS &\footnotesize MGS & \scriptsize{CW\!$\times$\!M} \\
       &  &      &       &   &  &   \scriptsize E-NC  & \footnotesize -NC &   \footnotesize -5NN &\footnotesize -1NN &\footnotesize -GRF & \scriptsize{GS-GRF}  \\
     \midrule
     {\small Pr-at} 
      
     & \footnotesize Churn  &0.894	&0.870	&0.847	&0.632	&0.850	&0.908	&0.910	&0.913 &\textbf{0.930}&- \\ 
     {\small -rec}
      & \footnotesize Mark. &0.119	&0.118	&0.115	&0.106	&0.093	&0.128	&0.126	&0.128 &\textbf{0.129} &-\\
     ($0.2$) & \footnotesize Private &\textit{Ref.}   &\textit{+9}\%  &+2\% & \textit{+9}\% &-34\% & +7\%  &\textit{+9}\% &\textit{+9} \%  &\textit{+9}\% &\textbf{+13}\% \\

     \midrule
     \multirow{2}{*}{\small PR }  
     & Churn  &0.622	&0.608	&0.576	&0.394	&0.595	&0.655	&0.653	&0.663 &\textbf{0.664} &- \\
     \multirow{2}{*}{\small AUC} 
     & \footnotesize Mark. &0.092	&0.090	&0.090	&0.082	&0.076	&0.099	&0.099	&0.098 &\textbf{0.100} &- \\
     & \footnotesize Private &\textit{Ref.}     &\textit{+11}\%  &+7\%  &+10\% &-28\%,  &+8\%   &+8\%,  &+10\%,  &\textit{+11}\%	&\textbf{+15}\%\\
     
     \midrule
     \multirow{2}{*}{\small ROC}  
     & \footnotesize Churn  &0.977	&0.971	&0.963	&0.941	&0.975	&0.983	&0.983	&\textbf{0.984} &\textbf{0.984} &- \\
     \multirow{2}{*}{\small AUC}
     & \footnotesize Mark. &0.890	&0.882	&0.878	&0.881	&0.861	&\textbf{0.899} &\textbf{0.899} &\textbf{0.899} &{0.898} &- \\   
     & \footnotesize Private  &\textit{Ref.}     &+0\%     &+0\%     &+0\%    &-2\%   &+0\%     &+0\%    & +0\%     &+0\%   &+0\%	\\
     \midrule
     {\small Time}  
     & \footnotesize Churn. &0.296	 &0.333	&0.494	&0.060	&0.852	&2.158   &1.245   &1.217   &0.893 &- \\ 
     (s) & \footnotesize Mark  &1.294	 &1.338	&1.919	&0.288	&4.214   &16.274   &8.050   &7.767  &5.869 &-\\
     \bottomrule
\end{tabular}
\end{table}

\subsection{Results}

In \Cref{tab:merged_results}, we first observe that oversampling strategies that preserve coherence (MGS-1NN and MGS-GRF) leads to better predictive performances than the non-coherent ones (SMOTE-NC, MGS-NC, MGS-5NN).
Besides, we remark that SMOTE-NC induces the greatest deterioration of predictive performances, 
for example $-28\%$ of PR AUC on the private data set. 
Furthermore, MGS-NC strategy leads to better predictive performances than SMOTE-NC for all three data sets, reinforcing conclusions of \cite{sakhowe}: the MGS KDE is better suited than SMOTE linear interpolation for minority class continuous features regeneration.
We also see that our proposed method MGS-GRF has the best predictive performances for BankChurners and BankMarketing in \Cref{tab:merged_results}
for all metrics, with a running time (for oversampling and LightGBM training) close to that of SMOTE-NC.

For the private data set, we observe that MGS-GRF and CW are the two best strategies (in italics). Those are promising results for our method and validate our findings of \Cref{subsec:sim2} and \Cref{subsec:sim-high-dim} on association and coherence. To take advantage of both strategies, we built an ensemble learning model CW$\times$MGS-GRF combining the two LightGBM obtained after CW and MGS-GRF strategies. This methodology obtains the best results by far. 



%% file: conclusion.tex
\section{Conclusion and perspectives}

In this paper, we propose an oversampling strategy, MGS-GRF which synthesizes continuous features with a kernel density estimator and categorical ones by a GRF.
%
%
We show through our first experimental protocol  with simulated data (\Cref{subsec:sim2}), that coherent strategies (MGS-1NN and MGS-GRF) lead to better predictive performances in terms of PR AUC and ROC AUC.
Then, in \Cref{subsec:sim-high-dim}, we show that nearest-neighbor based oversampling strategies are not well suited to handle categorical variables, since they do not preserve the association of generated samples in the presence of noisy features. 

We performed numerical experiments on two real-world open source data sets and an industrial private data set from a financial institution which is used in production. Our results show that MGS-GRF is the most promising strategy for real-world applications, achieving the best predictive performances. Thus, we recommend designing strategies are coherent and preserve association and, among those, we recommend the use of our proposed method MGS-GRF, which achieves the best predictive performances on real-world data sets.




%% file: appendix.tex

\appendix
\section{Supplementary materials}
\label{app:supp-mat}

\begin{figure}
    \centering
    \includegraphics[width=0.9\textwidth]{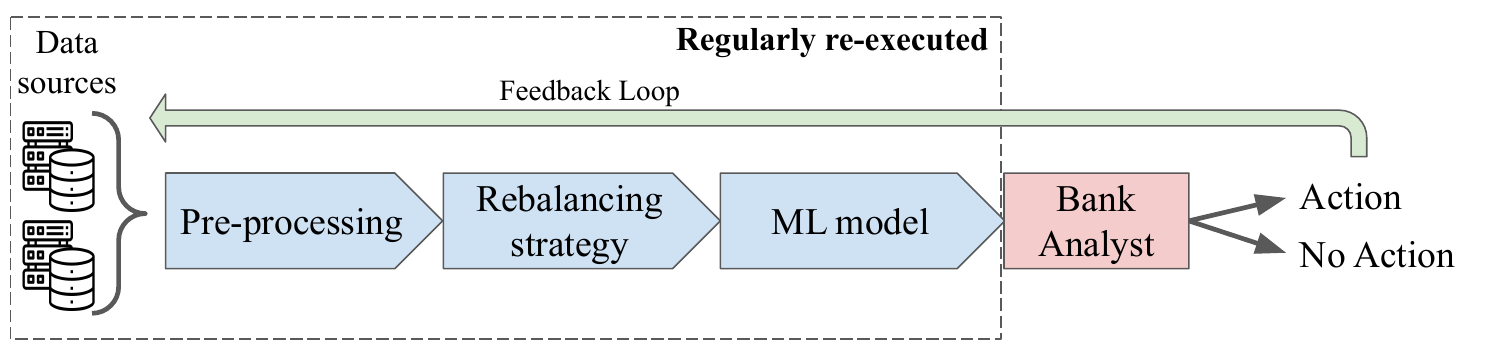}
    \caption{Pipeline of private data set described in \Cref{sec:exp-data-real}.}
    \label{fig:pipeline}
\end{figure}

\begin{table}
\centering
\caption{\Cref{tab:sim2-3-corps} with standard deviations. }
\label{tab:sim2-3-annexe}
\setlength\tabcolsep{2.5pt}
\begin{tabular}{llllllllll}
     \toprule 
       Strategy &  None    &  CW     &  ROS & RUS &   \small SMOTE  &MGS&   MGS &MGS   &    MGS \\ 
       &         &  & & & -NC  &  -NC & -kNN &-1NN   &-GRF \\ 
     \midrule
     \small PR AUC &	0.903	&0.903	&0.893	&0.699	&0.860	&0.922	&0.870	&0.952 &\textbf{0.954} \\
     \tiny std & \tiny$\pm0.054$ 	& \tiny$\pm$0.055	& \tiny$\pm$0.056	& \tiny$\pm$0.121	& \tiny$\pm$0.041	& \tiny$\pm$0.030	& \tiny$\pm$0.045	& \tiny$\pm$0.025 & \tiny$\pm$0.023 \\
     \small ROC AUC   &0.975	&0.977	&0.975	&0.935	&0.962	&0.984	&0.970	&\textbf{0.993}	&\textbf{0.993}\\
     \tiny std & \tiny$\pm$0.014	& \tiny$\pm$0.012	& \tiny$\pm$0.014	& \tiny$\pm$0.030	& \tiny$\pm$0.012	& \tiny$\pm$0.008	& \tiny$\pm$0.013	& \tiny$\pm$0.005	& \tiny$\pm$0.005 \\
     $COH$ & \textbf{100}\%	& \textbf{100}\% & \textbf{100}\% &\textbf{100}\%	& 90\%	& 90\%	&	83\% &	\textbf{100}\% &\textbf{100}\%	\\ 
     \tiny std & \tiny$\pm0$ & \tiny$\pm0$ & \tiny$\pm0$ & \tiny$\pm0$ & \tiny$\pm1$ & \tiny$\pm1$ & \tiny$\pm1$ & \tiny$\pm0$ & \tiny$\pm0$ \\
     Time (s) &0.55	&0.56	&0.74	&0.27	&1.01  &1.23  &1.04   &1.00   &\textbf{1.43} \\
     \tiny std &\tiny$\pm$0.03	&\tiny$\pm$0.03	&\tiny$\pm$0.04	&\tiny$\pm$0.01	&\tiny$\pm$0.02	&\tiny$\pm$0.01  &\tiny$\pm$0.01  &\tiny$\pm$0.04  &\tiny$\pm$0.05 \\
     \bottomrule
\end{tabular}
\end{table}

\begin{table}
\centering
\caption{\Cref{tab:merged_results} with standard deviations. For confidentiality motivations, all metrics of the private data set are relative gains compared to None strategy.}
\label{tab:merged_results_std}
\setlength\tabcolsep{2pt}
\begin{tabular}{llllllllllll}
     \toprule 
      \multirow{2}{*}{\footnotesize Metric} & \multirow{2}{*}{\footnotesize Data} & \multicolumn{9}{c}{Strategy} \\
      \cmidrule(lr){3-12}
       &  &  \footnotesize None    &  \footnotesize CW     &  \footnotesize ROS & \footnotesize RUS &   \scriptsize SMOT  & \footnotesize MGS &   \footnotesize MGS & \footnotesize MGS &\footnotesize MGS & \scriptsize{CW $\times$} \\
       &  &      &       &   &  &   \scriptsize E-NC  & \footnotesize -NC &   \footnotesize -5NN &\footnotesize -1NN &\footnotesize -GRF & \scriptsize{M-GRF}  \\
     \midrule
     {\small Pr-at} 
      
     & \footnotesize Churn  &0.894	&0.870	&0.847	&0.632	&0.850	&0.908	&0.910	&0.913 &\textbf{0.930}&- \\ 
     {\small -rec}
     &           \tiny std  & \tiny$\pm0.051$ & \tiny$\pm0.056$ & \tiny$\pm0.061$ & \tiny$\pm0.123$ & \tiny$\pm0.053$ & \tiny$\pm0.057$ & \tiny$\pm0.048$ & \tiny$\pm0.050$ & \tiny$\pm0.052$ &- \\
      & \footnotesize Mark. &0.119	&0.118	&0.115	&0.106	&0.093	&0.128	&0.126	&0.128 &\textbf{0.129} &-\\
     &           \tiny std & \tiny$\pm0.009$ & \tiny$\pm0.009$ & \tiny$\pm0.011$ & \tiny$\pm0.012$ & \tiny$\pm0.007$ & \tiny$\pm0.014$ & \tiny$\pm0.008$ & \tiny$\pm0.012$ & \tiny$\pm0.008$ &- \\
     ($0.2$) & \footnotesize Private &\textit{Ref.}   &\textit{+9}\%  &+2\% & \textit{+9}\% &-34\% & +7\%  &\textit{+9}\% &\textit{+9} \%  &\textit{+9}\% &\textbf{+13}\% \\

     \midrule
     \multirow{2}{*}{\small PR }  
     & Churn  &0.622	&0.608	&0.576	&0.394	&0.595	&0.655	&0.653	&0.663 &\textbf{0.664} &- \\
     \multirow{2}{*}{\small AUC} 
     &           \tiny std  & \tiny$\pm0.024$ & \tiny$\pm0.026$ & \tiny$\pm0.024$ & \tiny$\pm0.043$ & \tiny$\pm0.021$ & \tiny$\pm0.026$ & \tiny$\pm0.024$ & \tiny$\pm0.022$ & \tiny$\pm0.025$ &- \\
     & \footnotesize Mark. &0.092	&0.090	&0.090	&0.082	&0.076	&0.099	&0.099	&0.098 &\textbf{0.100} &- \\
     \multirow{2}{*}{\tiny std}  
     &           \tiny std & \tiny$\pm0.008$ & \tiny$\pm0.005$ & \tiny$\pm0.006$ & \tiny$\pm0.006$ & \tiny$\pm0.004$ & \tiny$\pm0.005$ & \tiny$\pm0.005$ & \tiny$\pm0.006$ & \tiny$\pm0.006$ &- \\
     & \footnotesize Private &\textit{Ref.}     &\textit{+11}\%  &+7\%  &+10\% &-28\%,  &+8\%   &+8\%,  &+10\%,  &\textit{+11}\%	&\textbf{+15}\%\\
     
     \midrule
     \multirow{2}{*}{\small ROC}  
     & \footnotesize Churn  &0.977	&0.971	&0.963	&0.941	&0.975	&0.983	&0.983	&\textbf{0.984} &\textbf{0.984} &- \\
     \multirow{2}{*}{\small AUC}
     &           \tiny std  & \tiny$\pm0.005$ & \tiny$\pm0.005$ & \tiny$\pm0.006$ & \tiny$\pm0.008$ & \tiny$\pm0.004$ & \tiny$\pm0.002$ & \tiny$\pm0.003$ & \tiny$\pm0.003$ & \tiny$\pm0.002$ &- \\
     & \footnotesize Mark. &0.890	&0.882	&0.878	&0.881	&0.861	&\textbf{0.899} &\textbf{0.899} &\textbf{0.899} &{0.898} &- \\   
     \multirow{2}{*}{\tiny std}      
     &           \tiny std & \tiny$\pm0.003$ & \tiny$\pm0.004$ & \tiny$\pm0.003$ & \tiny$\pm0.004$ & \tiny$\pm0.004$ & \tiny$\pm0.003$ & \tiny$\pm0.002$ & \tiny$\pm0.003$ & \tiny$\pm0.003$ &- \\
     & \footnotesize Private  &\textit{Ref.}     &+0\%     &+0\%     &+0\%    &-2\%   &+0\%     &+0\%    & +0\%     &+0\%   &+0\%	\\
     \midrule
     {\small Time}  
     & \footnotesize Churn. &0.296	 &0.333	&0.494	&0.060	&0.852	&2.158   &1.245   &1.217   &0.893 &- \\ 
     &    \tiny std & \tiny$\pm$0.015	& \tiny$\pm$0.022	& \tiny$\pm$0.032	& \tiny$\pm$0.003	& \tiny$\pm$0.069	& \tiny$\pm$0.051  & \tiny$\pm$0.038   & \tiny$\pm$0.056   & \tiny$\pm$0.033 &- \\
     (s) & \footnotesize Mark  &1.294	 &1.338	&1.919	&0.288	&4.214   &16.274   &8.050   &7.767  &5.869 &-\\
     &    \tiny std & \tiny$\pm$0.022	& \tiny$\pm$0.033	  & \tiny$\pm$0.088	  & \tiny$\pm$0.010	  & \tiny$\pm$0.054   & \tiny$\pm$0.563  & \tiny$\pm$0.086   & \tiny$\pm$0.203  & \tiny$\pm$1.342 &- \\
     \bottomrule
\end{tabular}
\end{table}

\clearpage
\section{Details on protocols}
In this section we give several details on our numerical experiments.

\subsection{Numerical illustrations of non-coherence phenomenon}
The protocol from \Cref{subsec:sim2} with 6 configuration values for $\Theta,\alpha,\Gamma$. For each configuration, $50$ different data sets (with different seeds) are generated. All in all, we obtain $300$ datasets. Finally, each data set is composed of $5000$ samples with an imbalance ratio less than $10\%$. 

For each data set, we apply different rebalancing strategies and apply a LightGBM classifier on the rebalanced data set.

\subsection{Protocol : studying neighborhood based strategies in high dimensional setting}
\label{sec:app-sim-high-dim}

The protocol from \Cref{subsec:sim-high-dim} is executed with the following dimensions values $d$ : $[5,10,20,30,50,100,150,200]$. All the dimensions share the same parameter values. Each dimension simulation is executed $20$ times in order to be able to compute standard deviations.

Regarding the generation of the samples, let \(\pi_1, \pi_2, \pi_3\) be the proportions of these three Gaussians, we have \(\pi_1 = \pi_2 \gg \pi_3\).

\paragraph{Details on \Cref{fig:sim-high-dim-asso}}

The Bayes Classifer for the $Asso$ of \Cref{fig:sim-high-dim-asso} is derived empirically from a LightGBM trained on continuous features to predict the categorical feature. This latter model is trained on a different simulated data sets with millions of samples. We make this choice because we predict only one categorical feature and because LightGBm is a consistent estimator.